\documentclass[lettersize,journal]{IEEEtran}
\usepackage{amsmath,amsfonts}
\usepackage{algorithmic}
\usepackage{algorithm}
\usepackage{array}
\usepackage[caption=false,font=normalsize,labelfont=sf,textfont=sf]{subfig}
\usepackage{textcomp}
\usepackage{stfloats}
\usepackage{url}
\usepackage{verbatim}
\usepackage{graphicx}
\usepackage{cite}
\begin{document}

\title{Reconstructing High-Dimensional Datasets From Their Bivariate Projections}

\author{Eli Dugan, Klaus Mueller,~\IEEEmembership{Fellow,~IEEE}
\thanks{Eli Dugan is with Williams College, Williamstown, MA and Klaus Mueller is with Stony Brook University, Stony Brook, NY. The research was done when the first author was a NSF REU fellow at Stony Brook University. E-mails: ebd3@williams.edu; mueller@cs.stonybrook.edu.}
\thanks{Manuscript received May 5, 2023.}}



\maketitle

\begin{abstract}
This paper deals with developing techniques for the reconstruction of high-dimensional datasets given each bivariate projection, as would be found in a matrix scatterplot. A graph-based solution is introduced, involving clique-finding, providing a set of possible rows that might make up the original dataset. Complications are discussed, including cases where phantom cliques are found, as well as cases where an exact solution is impossible. Additional methods are shown, with some dealing with fully deducing rows and others dealing with having to creatively produce methods that find some possibilities to be more likely than others. Results show that these methods are highly successful in recreating a significant portion of the original dataset in many cases - for randomly generated and real-world datasets - with the factors leading to a greater rate of failure being lower dimension, higher n, and lower interval.
\end{abstract}


\section{Introduction}

\begin{figure}[htbp]
\centering
\includegraphics[width=3.4in]{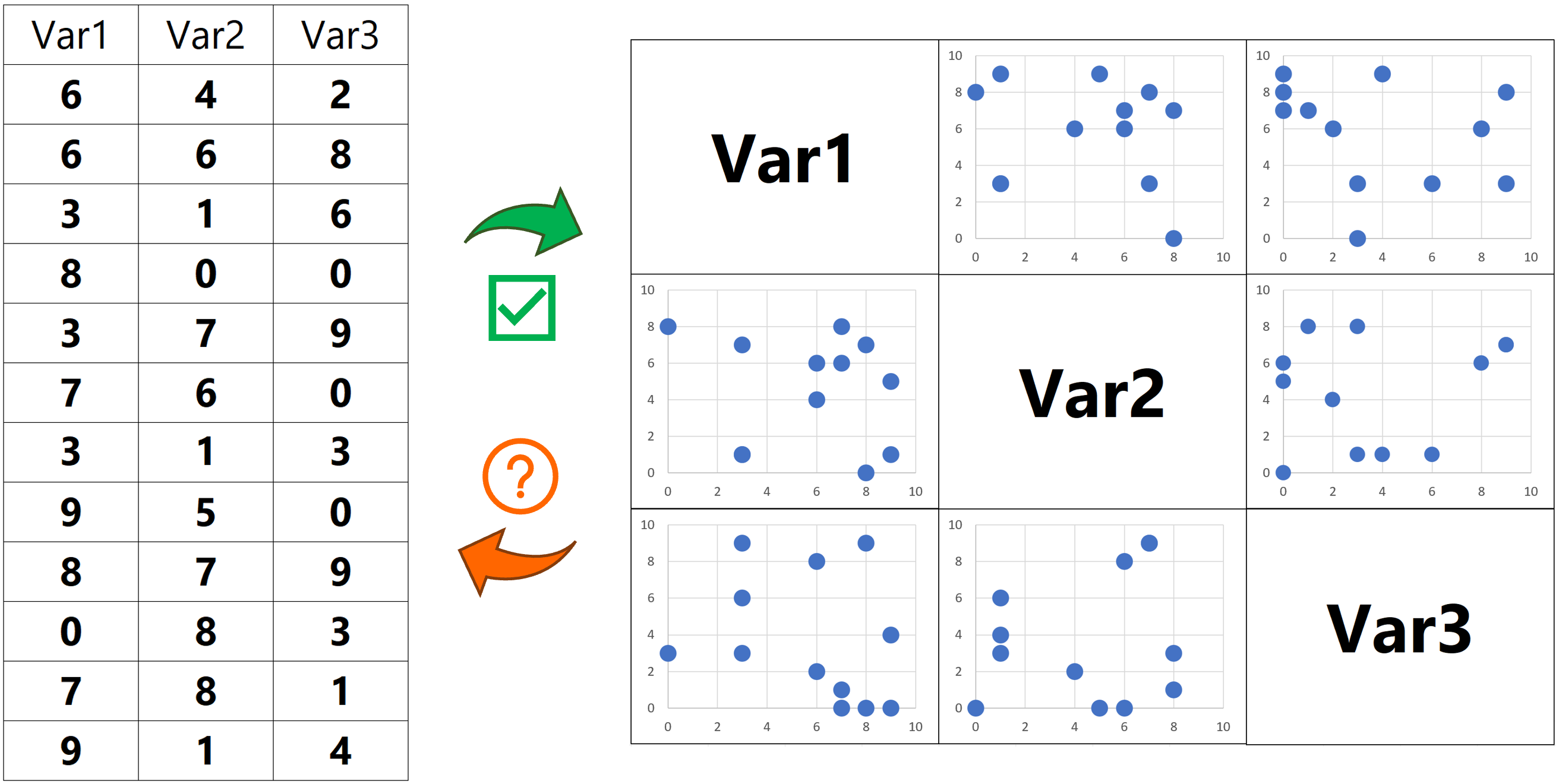}
\caption{The question statement}
\label{scatterplotmatrices}
\end{figure}

\IEEEPARstart{A}{ny} dataset having three or more columns, with entries being numeric values, can be visualized using a scatterplot matrix. Each possible pair of variables (a bivariate projection of the initial dataset) is represented by a scatterplot, where each point in each scatterplot represents a coordinate pair of values from the corresponding pair of variables. This paper asks the question of whether it is possible to use the information that can be found in a scatterplot matrix – every pair of values within every pair of variables – to determine what the original dataset must have been. An example of this question is pictured in Fig. \ref{scatterplotmatrices}.

This means that the information given as an input to the methods can be thought of as a collection of all possible pairs of columns (bivariate projections) from the original dataset. This allows us to see duplicate coordinate pairs that may come from distinct rows. Note that we cannot assume that these projections are ordered in the same way that they are in the original dataset. Also, note that we are studying the reconstruction of datasets with projections that have a dimension of two, but it is possible to generalize some of these methods to higher dimensional projections.

For any given dataset, the term “dimension” will refer to the number of variables it has, which is equivalent to the number of columns it has, or its dimensionality. Also, “n” will refer to the number of entries, or rows, in the dataset. Each column has an “interval,” which refers to the number of possible values there are for that column. This paper will mostly deal with randomly generated and integer-valued datasets where each column has the same interval (such that the whole dataset can be said to have a uniform interval). However, the methods discussed here generalize beyond each of these given restrictions. It is possible to apply these methods to datasets with non-integer values, and even non-numeric values. Real-world (nonrandom) datasets will be tested as well.

\section{Related Work}

The reconstruction of numerical data from their projections has been researched for decades, and many algorithms have been proposed. They depend on the nature of the data the projections originate from. We distinguish these as follows.

If the original domain is a continuous 2D image or a 3D volumetric object, then the projections are generated by integrating the object’s values (densities) along linear rays. This is called the \textit{Radon Transform} and analogously, the reconstruction of a 2D image or 3D volume from its projections is called the \textit{inverse Radon Transform}. It is the type of process that occurs in X-ray imaging with Computed Tomography (CT) \cite{herman2009fundamentals}. Many CT reconstruction algorithms have been proposed in the past, such as the Algebraic Reconstruction Technique (ART) \cite{ gordon1970algebraic}, Simultaneous ART (SART) \cite{andersen1984simultaneous}, Multiplicative ART (MART) \cite{gordon1970algebraic}, the Simultaneous Iterative Reconstruction Technique (SIRT) \cite{ gilbert1972iterative}, and Expectation Maximization (EM) \cite{lange1984reconstruction}. All of these techniques use numerical optimization to accomplish the task since a direct Inverse Radon Transform is usually not possible due to noise in the image generation process. Since these algorithms can be time-intensive, especially with large data and high-resolution targets, methods to accelerate these operations on the GPU have been devised \cite{xu2005accelerating}. More recently, learning-based approaches have been able to learn the projection operators for a variety of data scenarios and so achieve significant speed-ups (see survey by Wang et al. \cite{wang2018image}). 

Conversely, the reconstruction of a point cloud from its projections requires the solution of a correspondence problem. Reconstructing a set of 3D points from a pair of stereoscopic projections is called the \textit{Epipolar Transform}. It requires that a 3D point to be reconstructed be  distinguishable in each 2D projection. Recent algorithms relax this constraint by using learning-based feature analysis to use feature information from multiple views to resolve ambiguities in 3D (see e.g. \cite{ he2020epipolar}). A common method to acquire point-based data of 3D objects for later surface reconstruction is via RGB cameras with depth sensors, giving rise to RGB-D data. The reconstruction typically uses deep neural networks (see survey \cite{ liu2019deep}), but for the most part, these algorithms operate in 3D space and seek to reconstruct exterior surfaces. 

Our work is more general in that it extends the problem into higher dimensions, such as 6D and more. The point clouds here are (volumetric) data distributions and not surfaces. While these distributions could have manifolds, the projections hold all points and not just those on the manifold. In addition, the projections are not ray integrals, and they do not have depth information. Further, the point cloud in high-D is often sparse and irregularly distributed, although the points might follow some statistical description, such as clusters or a trend. Hence, the overall problem does not directly fit the two situations outlined above, yet there has been some work in this area. Iterative proportional fitting (IPF) \cite{deming1940least} revises a joint distribution that is close to the true solution with a set of marginal distributions. But IPF cannot always find a preferred solution with insufficient information as it uses a similar process as MART. Statistical methods use GMM \cite{romeo2005estimating} or Bayesian approaches \cite{chaubey2003estimation} \cite{putler1996bayesian} to estimate the real joint distribution based on a prior joint distribution from smaller subsets of data. However, the prior distribution is unavailable in most cases. Some sampling methods \cite{chen2005lattice} \cite{dyer1997sampling} \cite{kannan1997sampling} reconstruct contingency tables from the marginal frequencies by sampling the lattice points in high-dimensional space, but these techniques are not scalable to large datasets. To address the issue that the reconstruction of multivariate data from their lower-dimensional projections is often ill-posed, Xie et al. \cite{xie2016visual} developed an interactive visual approach where humans could use their domain understanding to add constraints to narrow the solution space. Our approach seeks to tackle the problem with more sophisticated analytics; remaining ambiguities could be resolved via an approach similar to that of Xie et al.

\section{Lookup Method}

\begin{figure}[htbp]
\centering
\includegraphics[width=3.4in]{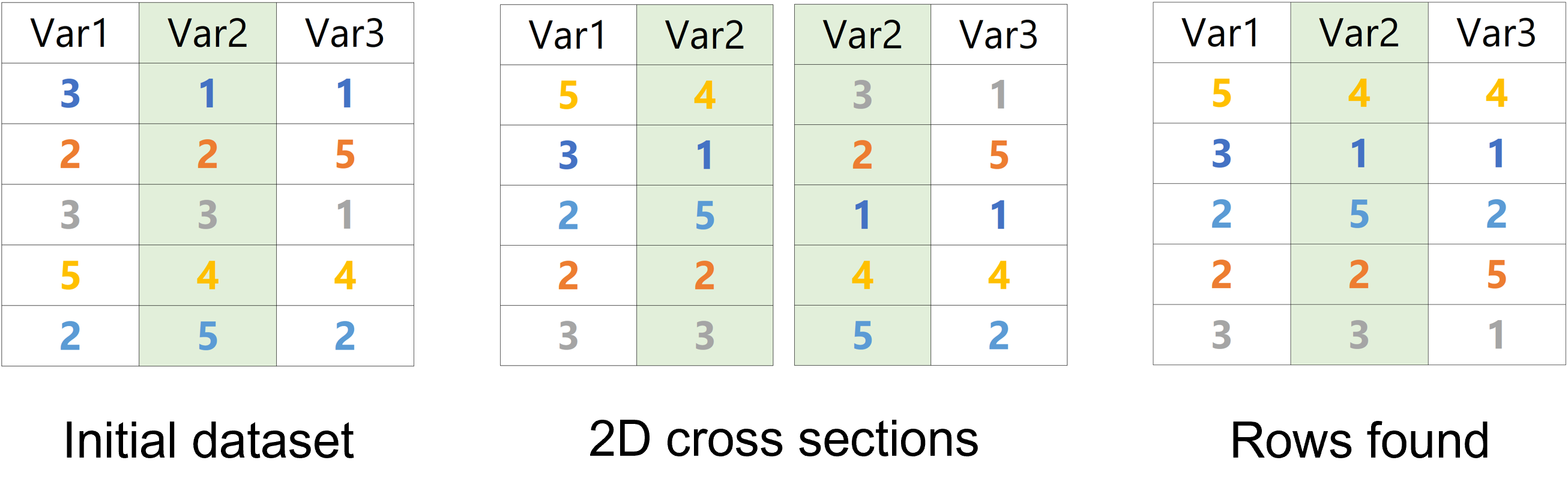}
\caption{The Lookup Method}
\label{lookupmethod}
\end{figure}

We can first consider a method that recovers the entirety of a dataset in a subset of cases. In particular, the following method can be used if there exists at least one column in the original dataset that contains no duplicate values. If such a column (one with all values distinct) is identified, this “Lookup Method" (pictured in Fig. \ref{lookupmethod}) uses the bivariate projections that include that column and one other (so, the method uses one fewer projection than the dimension of the dataset). Since each value in the identified column exists only once, it will exist only once in each of the projections being used, and will be in only one coordinate pair with a different value of a different variable. For a given unique value, the values from each of its associated coordinate pairs will comprise one original row of the dataset, and this process can be repeated for each of these values to recreate the full dataset.

For a given set of initial parameters known about the dataset, it is possible to find the probability that the method will be successful. This is done by calculating the chance that at least one column will have no duplicates. For any individual column with n of $N$ and an interval of $I$, the chance it will have no duplicates is the number of ways to choose $N$ items from $I$ possibilities without replacement divided by the total number of ways to choose $N$ items from $I$ possibilities with replacement, which is given by

\begin{equation}
{I \choose N}I ^ {-N}.
\end{equation}

Then, the chance of the method failing overall is the chance that each column fails independently. For a dataset with $D$ dimensions, and with $I_d$ being the interval of the $d\textsuperscript{th}$ dimension, this is given by

\begin{equation}
\prod_{d=1}^{D} \left( 1-{I_d \choose N}I_d ^ {-N} \right).
\end{equation}

If no column with all distinct values exists in the dataset, then the Lookup Method will fail. This happens because, for some non-distinct value, there will be multiple possible coordinate pairs within a given projection that include that value, which means that original rows cannot always be precisely recovered through this method alone. Note that if a dataset’s n is greater than its interval, the method is guaranteed to fail, since a duplicate is guaranteed by the Pigeonhole Principle. So, more complex methods are needed to solve the cases of datasets where every column has duplicate values.

\section{Graphs}

\subsection{Clique-finding}

\begin{figure}[htbp]
\centering
\includegraphics[width=3.4in]{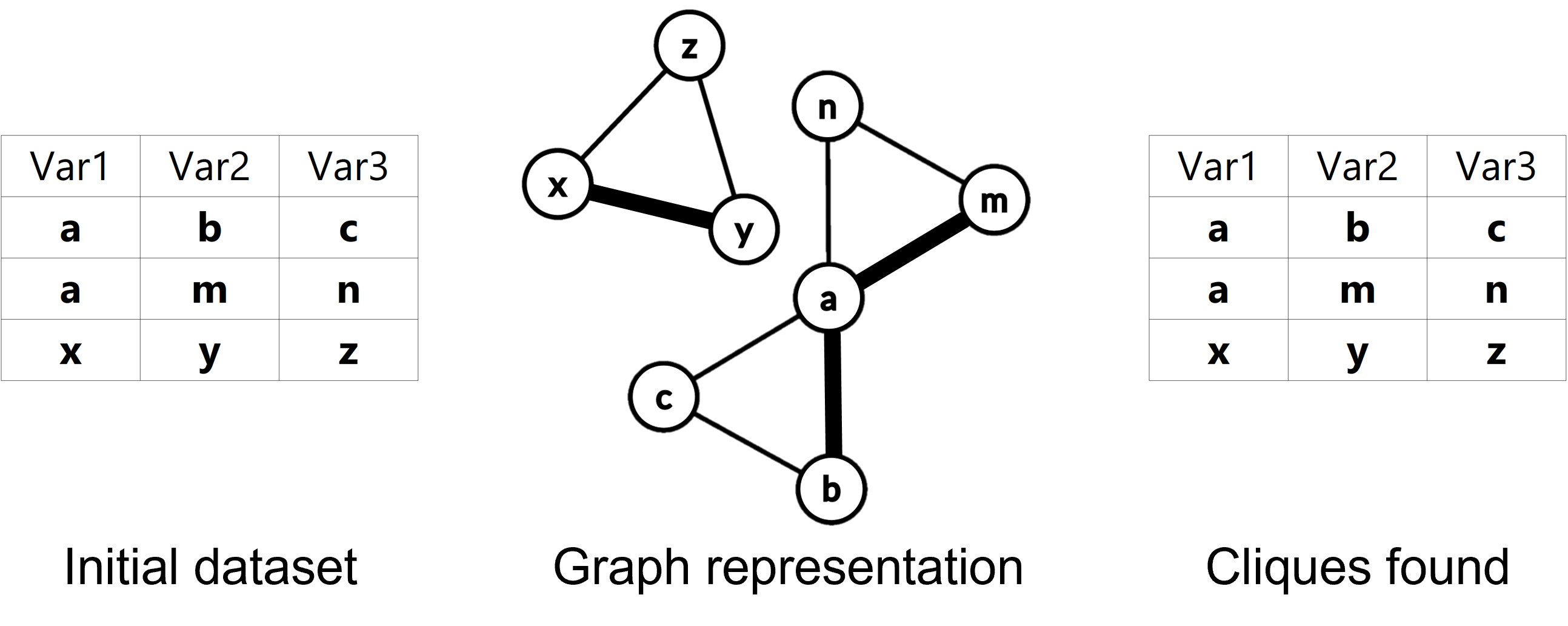}
\caption{Clique-finding example}
\label{cliques}
\end{figure}

Having been given every bivariate projection of the initial dataset, we know every value that exists within each column. Suppose we create a graph in which every value of each variable is represented by a vertex. Note that if the same value appears in more than one variable, separate vertices are created for each instance, but multiple instances of one value within a variable are represented by a single vertex. For every coordinate pair in each projection, suppose we create an edge between the two vertices in our graph relating to these two values. This is accomplished by storing information about which variable the vertex belongs to, as well as the value itself, in the vertex. For example, as is seen in Fig. \ref{cliques}, the coordinate pairs for variables 1 and 2 are bolded.

For any subset of the resulting graph that includes only values from a single row of the original dataset, every vertex will be connected by an edge to every other vertex. This is because each pair of values within any row is represented as a coordinate pair within that corresponding bivariate projection of the whole dataset, which means that that connection is seen as an edge in the graph. So, we have that a row of the original dataset will manifest as a fully connected subcomponent of the original graph – a clique. Specifically, the number of vertices in the clique will be equal to the dimension of the original dataset. Having found such a clique, we can determine what that original row was by looking at each of the values of each vertex, along with which variable they belong to.

For any graph representation of the bivariate projections of some dataset, then, it is possible to run a search for all k-cliques in that graph (where k is the dimension of the dataset). Since every row in the dataset is guaranteed to be a clique in the graph, if the number of cliques found in the graph representation is equal to the known number of distinct rows in the dataset, then the found cliques represent precisely that original dataset and can recreate it exactly. Unfortunately, this does not occur in every case, and the reasons why this is the case are discussed in section \ref{phantomsection}.

While there is no direct formula for the success rate of this unembellished clique method at this time, we can conduct tests to determine that it is more powerful than the Lookup Method. Taking as an example the case of a dataset with dimensionality 6, n of 12, and a uniform interval of 10, we immediately know that the Lookup Method will fail, since n is greater than the interval. However, in 1000 simulations of random datasets fitting these parameters and simply creating the graph and finding cliques in the manner described above, a dataset created using the identified cliques was equivalent to the original dataset 86\% of the time, which is significantly higher than the 0\% from the Lookup Method.

\subsection{Phantoms} \label{phantomsection}

\begin{figure}[htbp]
\centering
\includegraphics[width=3.4in]{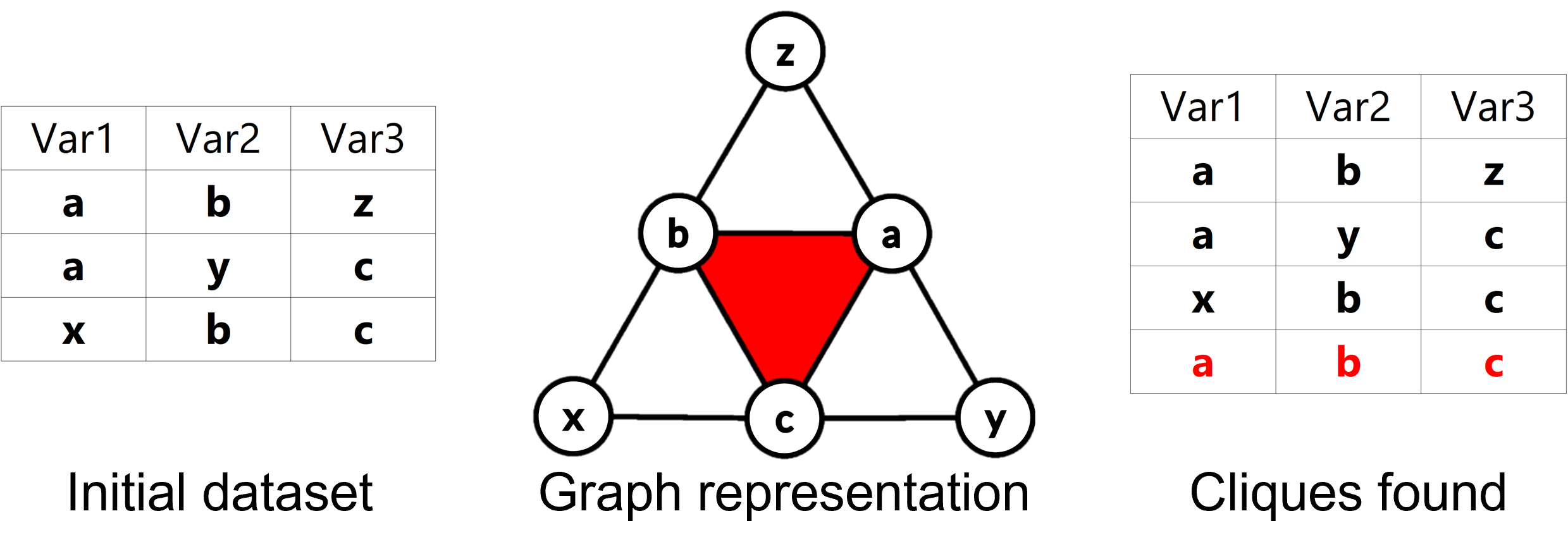}
\caption{Phantoms}
\label{phantoms}
\end{figure}

Unfortunately, even though it is true that every row in the original dataset will correspond to a clique in the graph representation, the converse is not always true: a clique in the graph representation does not always represent an original row. As was implied above, there can sometimes be more cliques in the graph than there are original rows, which means that a method that purely finds cliques of the correct size in the graph and reports the recreated dataset as having each of these as a row will sometimes be incorrect. We call these cliques that do not correspond to rows of the original dataset “phantoms,” an example of which is given in Fig. \ref{phantoms}.

As is demonstrated by Fig. \ref{phantoms}, phantoms are formed when coordinate pairs originating in different rows happen to form a clique. Each edge of the phantom exists somewhere in the original dataset, and sometimes a significant portion of a phantom will have originated from a small handful of similar rows. This is possible because the information that goes into the graph is purely the value pairs within the original dataset.

So, we have that the set of cliques found in our graph representation is the set of possible rows of the initial dataset. Every correct row will be contained in this set, but there may be phantoms that are not original rows. We will thus want to work out methods that can determine which of these cliques are real and which of them are phantoms.

\subsection{Impossible Cases}

It turns out that there exist cases that are not fully determinable, in that they cannot be recovered through their bivariate projections with certainty. These cases involve multiple different datasets that have the same edges within them, which means they have the same graph representation.

\begin{figure}[htbp]
\centering
\includegraphics[width=3.4in]{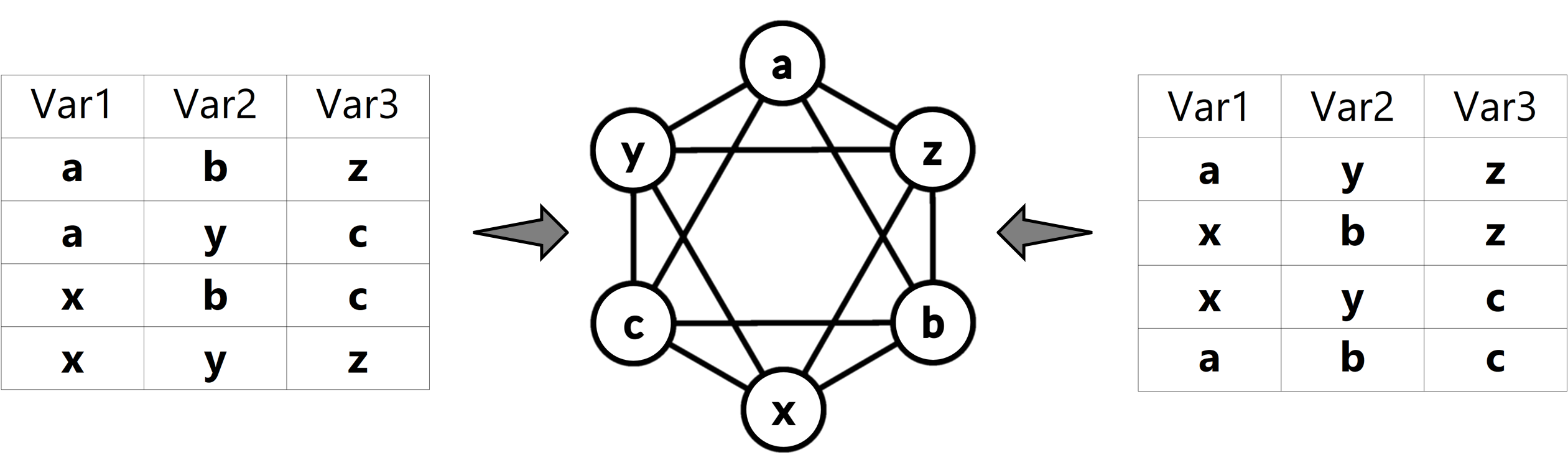}
\caption{A counterexample}
\label{impossiblecase}
\end{figure}

\begin{figure}[htbp]
\centering
\includegraphics[width=3.4in]{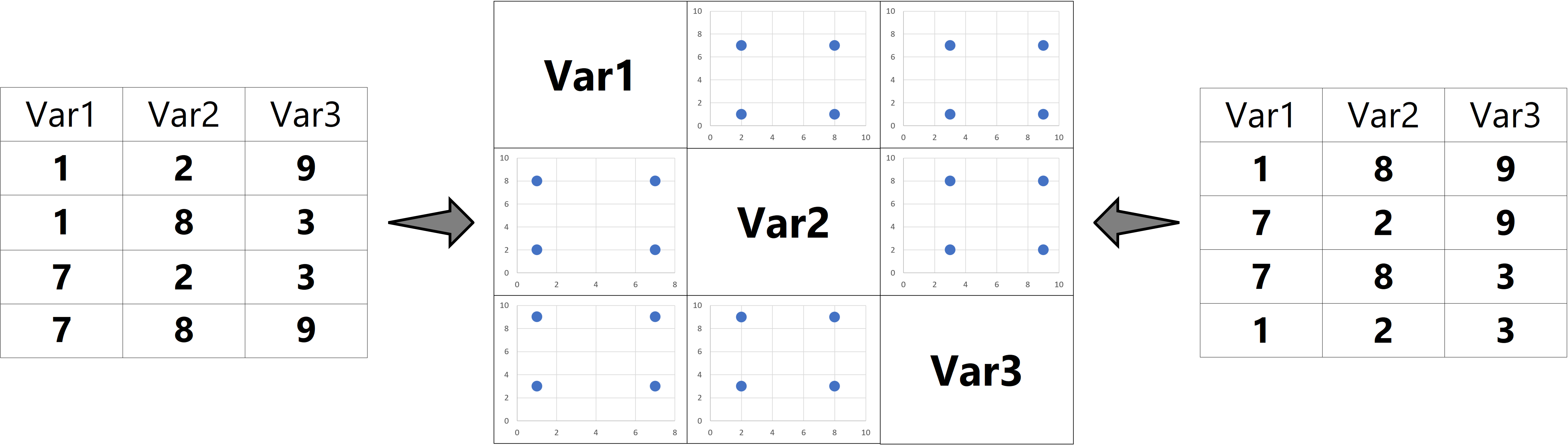}
\caption{Scatterplot matrix of counterexample}
\label{counterscatter}
\end{figure}

If it was the case that there was some way to always tell the original dataset from its graph representation, then we would expect that the relationship between graphs and datasets would be one-to-one - that is, that for any graph we might create, we know that there is a particular dataset that we conclude created that graph. If there was a case where two distinct datasets generated the same graph, the relationship would no longer be one-to-one, and we would not be able to determine which dataset generated the graph from the graph alone. An example of this kind is given in Fig. \ref{impossiblecase}. So, by way of this counterexample, we see that no method can be guaranteed to be able to determine the dataset that originally generated any given such graph. This same counterexample is given in Fig. \ref{counterscatter} to help visualize why we encounter issues with this case.

This means that there is a fundamental limit to the extent to which full deductions can be made about which cliques are true and which are phantoms, and that, depending on the case, there might always be some degree of uncertainty and margin for error. However, even if there exist cases that are not fully resolvable with certainty, we could still try to determine methods that can select cliques and be correct at a rate better than naive guessing might allow. To accomplish this, we may at some level want to think about how to quantify the likelihood of these rows being real and of them being phantoms.

\section{Methods of Recovery}

\subsection{The Tuples Method}

First, we wish to determine methods that can make full deductions of whether a particular clique – a possible original row – is correct, or whether there is uncertainty about whether it is a phantom. After the method involving creating a graph representation of the bivariate projections and finding all cliques of the correct size is completed, we are left with a set of possible rows. The next step is to create a new graph of all of these found cliques. Note that this graph must allow for the possibility of multiple edges, since an edge that appears multiple times in different possible rows will appear multiple times in this new graph of cliques. This means that the graph of cliques will not necessarily be identical to the previous graph representation – and if there are phantoms, then it will necessarily be different.

The Lookup Method identifies a column that has all distinct values, and each of these values “looks up” each of their connections to determine the original row that that value was a part of. Even if no such column exists and it is not possible to apply the Lookup Method, we can still apply the same principle to recover individual rows at a time. The Singles Method allows us to say that any of the possible rows that contain at least one value that is unique in its respective column can be deduced to have been in the original dataset. In terms of the graph of cliques, this means that for any vertex that is a part of just one clique, the clique that it is a part of is necessarily original. In this sense, the Lookup Method can be viewed as a special case of the Singles Method in which the entire dataset can be reconstructed using singles that happen to all be in the same column.

The Singles Method alone can recover entire datasets by using a combination of distinct singles coming from different columns, which makes it more powerful than the Lookup Method. For example, in Fig. \ref{phantoms}, the Singles Method could be applied to identify the first three possible rows as being correct, since they each contain a value that is unique within their respective columns, despite each original column of the dataset having duplicate values.

\begin{figure}[htbp]
\centering
\includegraphics[width=3.4in]{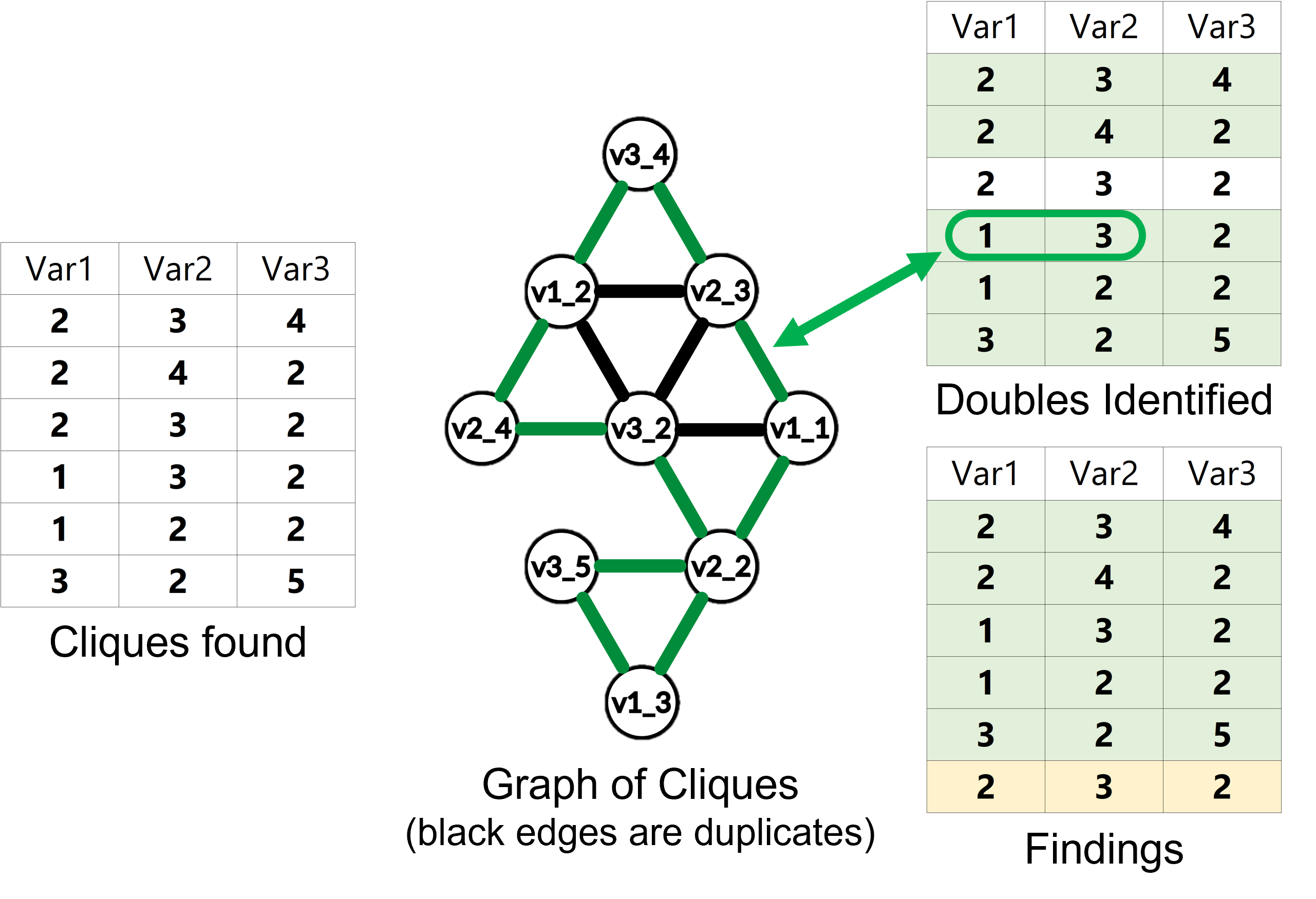}
\caption{The Doubles Method}
\label{doublesmethod}
\end{figure}

Instead of just looking at unique single values within their columns, we could instead determine which pairs of values within a row (doubles) are unique in their respective columns - that is to say, coordinate pairs that are unique within their bivariate projection. This observation allows us to generalize from the Singles Method to the Doubles Method, which is significantly more powerful. In this method, as is shown in Fig. \ref{doublesmethod}, we can say that any of the possible rows that contain any pair of values that are unique within their variables can be deduced to have been in the original dataset. When this logic is applied to the graph of cliques, it is equivalent to stating that every clique that has at least one non-shared edge necessarily represents an original row.

Note that the Doubles Method is strictly more powerful than the Singles Method. This is because if a single is flagged as unique, all the edges it is a vertex in will be flagged as being unique. So, the Doubles Method will necessarily be able to recover at least as many rows as the Singles Method. In one set of tests on random datasets with a dimension of 5, n of 100, and an interval of 25, the Singles Method correctly identified 11.8 rows on average, while the Doubles Method correctly identified 99.5 rows (practically the entire dataset, every time). For this reason, when implemented, only the Doubles Method is run.

The name “Tuples Method” comes from the idea that finding singles and doubles can be generalized to looking for unique instances of finite ordered lists of elements of some size. It turns out that, when taking bivariate projections, the 2-tuple (double) is the highest we can generalize, since that is the highest dimension of information given. However, we theorize that if these processes are repeated when the information given is a set of n-dimensional projections of some high-dimensional dataset, where n can be greater than 2, the Tuples Method would be able to generalize up to n-tuples to fully recover individual rows.

\subsection{Inferred Likelihood} \label{likelihood}

As was shown from the counterexample of the impossible case, we know that it is not always possible to determine with certainty the dataset that generated the bivariate projections that we are given. In such cases, we may be able to deduce that some of the possible rows were in the original dataset, through the Tuples Method, but we may not be able to recover as many rows as we know there are supposed to be, and we may be left with multiple options to choose between.

At this point, we know the entire set of edges (of coordinate pairs) that are in the original dataset, since this is what we were given with the bivariate projections. Every one of the cliques found in the original graph representation – every possible row – will be entirely composed of these edges that originated in the initial dataset. Some of these edges will be accounted for in the rows that the Tuples Method has fully recovered. However, not every one of the needed edges will necessarily be present in these deduced findings. Some of these needed edges may only exist within the set of remaining possible rows that have not yet been deduced. Note that if an edge exists only once in the set of cliques found, then it will be deduced to be original, assuming the Tuples Method has been fully applied. So, if an edge is not represented within the deduced findings, then there must be at least two possible rows within the set of rows that have not yet been deduced that contain that pair of values.

Due to the possibility of multiple rows in the initial dataset sharing a pair of values in some variable pair, more than one of these undeduced rows can be in the original dataset (note also that it could be the case that a possible row composed entirely of edges that have already been recovered is still original, though perhaps this is less likely). As such, for every unaccounted-for edge $e$ that exists in the initial dataset, we can generate a statement of the form “at least $x$ of $\{ r \in P \mid e \in r \}$ is correct." Here, $x$ is the required count of that edge in the original dataset, and $r$ is a row in $P$, the set of possible rows. The goal is now to determine a subset of the undeduced rows that satisfies all statements.

As an example, we might have a collection of possible rows \{A, B, C, D, E, F, G, H\}, of which three are correct, and three edges that are known to be in the dataset but have not yet been accounted for in the deduced rows: 1, 2, and 3, each of which only appears once in the initial dataset. Based on which of our possible rows include these edges, we might be able to construct the following statements:
\begin{enumerate}
  \item At least 1 of \{A, F, H\} is correct
  \item At least 1 of \{E, F, G\} is correct
  \item At least 1 of \{D, H\} is correct
\end{enumerate}
The task is then to select three of the possible rows such that all conditions are satisfied.

This is equivalent to a hard version of the Set Cover Problem, in which the task is to determine the smallest number of a given collection of subsets of some universal set that has a union equivalent to that universal set. In this case, we can think of all the unrecovered edges as being our universal set, and each subset corresponds to a possible row, where the elements in each subset are the edges in that row that do not appear in the deduced findings. Note that it is possible to reframe the entirety of this dataset reconstruction question in terms of Set Cover, but this will not be explored further at that level.

Given that we know how many rows are in the original dataset, and that we know how many rows we have recovered, the difference is the number of selections we can make to try to fill our best guess of the dataset (call this number of spaces available to make selections $s$). So, this reframed task is to pick $s$ of these subsets such that in their union, all these unrecovered edges are represented. It is possible to have multiple solutions, including possibly having solutions that require fewer than $s$ selections, and there is no known way of determining which of these options is the original. So, an optimal solution might generate all solutions of size $s$ that cover all needed edges and determine the proportion of the solutions where each possible row appears. These proportions could be interpreted as the likelihood that any given possible row is correct. The selection of the likeliest $s$ rows would then maximize the mean number of correct rows for any given case.

Returning to the previously given example, we may realize that there are multiple ways to choose three elements of \{A, B, C, D, E, F, G, H\} in a way that satisfies the aforementioned rules. As it turns out, there are 22 possibilities. Of these 22, the most common elements are H (15 of the 22 contain it), F (11), and D (10). Picking these three would therefore maximize the likelihood that we are getting these non-deduced rows correct.

Unfortunately, the optimal solution described above is highly computationally intensive and is infeasible even for datasets on the small side. After all, the regular Set Cover Problem is known to be NP-complete, and this optimal solution is harder still. As a result, an approximation is required. We also must make sure that we test against a naive guessing method to ensure that the results we get are significant. The method we found that seems to give the best results attempts to approximate the generated probabilities as explained above.

Each possible row will have a running total that will approximately correlate to a probabilistic estimate that it is original, and it relies on much of the same reasoning as the optimal Set Cover solution. For every statement of the form “at least $x$ of $\{ r \in P \mid e \in r \}$ is correct," where the cardinality of the set is $c$, then the value $\frac{1}{c}$ is added to the running total of each row in the set. Doing this for each unrecovered edge gives each possible row a corresponding value that correlates highly with the theoretical probability that the row was original. So, the rows corresponding to the highest $s$ values will be selected as the best guess of the rows that make up the original dataset. Note that $\frac{1}{c}$ is used as opposed to $\frac{x}{c}$; the latter was tried and did significantly improve the success rate.

Applying this method to the previously given example leads to the highest “scoring” elements being the same as those previously identified as most likely to be correct. Specifically, H had $\frac{1}{2} + \frac{1}{3} = \frac{5}{6}$, F had $\frac{1}{3} + \frac{1}{3} = \frac{2}{3}$, and D had $\frac{1}{2}$. These three were the highest scores of all 8 elements, so would be selected to give the best chance of getting these undeduced rows correct. Note that these values only have relative meaning to each other, and are not meant to be interpreted as probabilities.

\begin{figure}[htbp]
\centering
\includegraphics[width=3.4in]{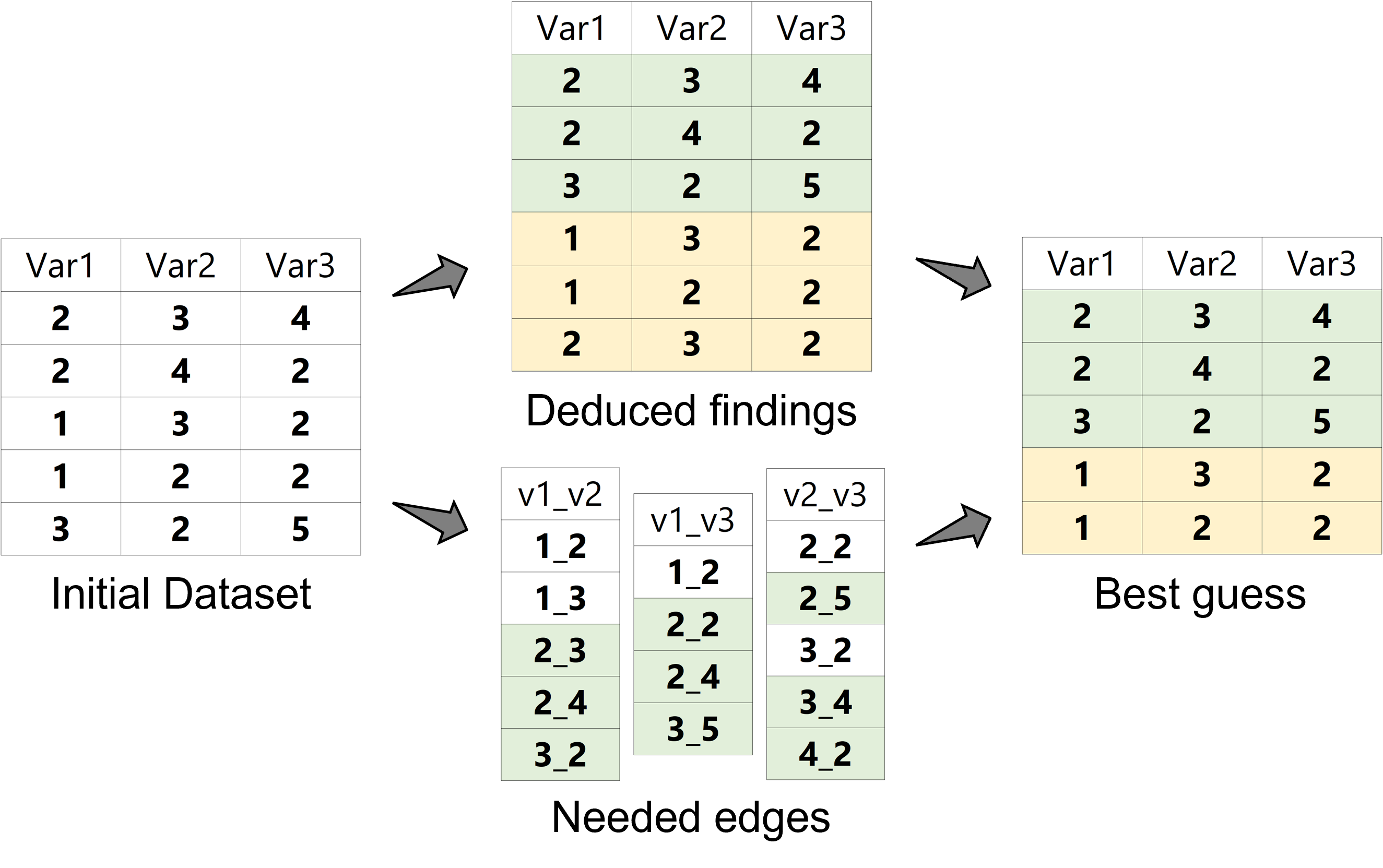}
\caption{Inferred Likelihood}
\label{inferredlikelihood}
\end{figure}

In Fig. \ref{inferredlikelihood}, we show an example of these methods applied to a sample dataset. Note that, for simplicity, the deduced findings are what would be deduced using just the Singles Method.

\section{Results}

\subsection{Principal Component Analysis Plots}  \label{pcasection}

\begin{figure}[htbp]
\centering
\includegraphics[width=3.4in]{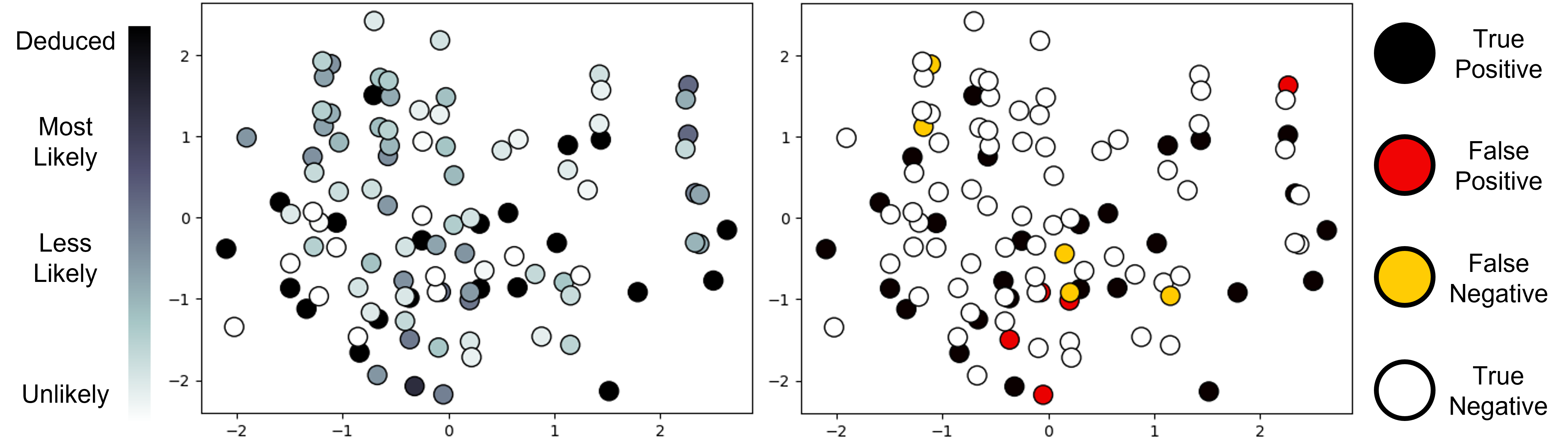}
\caption{Likelihood (left) and accuracy (right) PCA plots}
\label{pca}
\end{figure}

To visualize the inferred likelihood method, as well as to have a visual way to evaluate its success, we can display all the possible rows (found cliques) in a Principal Component Analysis (PCA) plot. Since each point represents a clique found, some of them represent original rows of the dataset and others represent phantoms. The example in Fig. \ref{pca} shows the results of a randomly generated dataset with a dimension of 5, n of 32, and a uniform interval of 8.

The left plot in Fig. \ref{pca} represents the “realness likelihood” of each of the possible rows, following the inferred likelihood method laid out in section \ref{likelihood}. The points in black are those that were deduced to be correct by the Tuples Method. All the other points went through the process of creating a probabilistic estimate of their inferred likelihood, and the “score” that each received based on how they best “fill the gaps” of the still-required coordinate pairs is shown by the gradient, where darker implies a higher score, and therefore a row that is more likely to be an original. In this example, 21 of the 32 rows were recovered by the Tuples Method, leaving 84 remaining options. So, to complete the guess, the next darkest 11 dots as represented in the plot were selected.

If we had picked 11 rows at random from these 84 options, the expected number of correct rows would be 1.4. In this example, 6 out of the 11 were correct. The plot on the right in Fig. \ref{pca} represents the accuracy of the guess, where the 6 undeduced rows that were correct (true positives) are now in black, in addition to the 21 fully deduced rows. The 5 incorrect guesses (false positives) are in red, while the 5 correct rows that were not guessed (false negatives) are in yellow. Finally, phantoms that were correctly identified as phantoms (true negatives) are colored in white.

\subsection{Results on Random Data}

There exist several possibilities by which one might evaluate the success of these metrics. Starting with judging the Tuples Method, the highest bar for success on a random dataset for some given set of parameters would be the proportion of the time that the entire dataset will be deduced. Another less strict method would be to find the average proportion of the distinct initial rows that are recovered by the Tuples Method. This second metric will always be greater than or equal to the first, as giving credit for any rows recovered will never be less than the credit given just for 100\% completion. As we know, there will always be cases that cannot guarantee 100\% accuracy.

These metrics can also be adapted for when the methods of inferred likelihood are applied after the Tuples Method. Since these methods do not guarantee 100\% recovery, it makes the most sense to evaluate these methods purely by looking at the average proportion of the dataset that will be recovered for some random dataset of given parameters. This will always be greater than or equal to the same proportion calculated for a random dataset with the same parameters with only the Tuples Method applied.  Again, due to the existence of impossible cases, this metric cannot be guaranteed to ever be 100\%.

\begin{table}[htbp]
\caption{Results table of random datasets}
\label{simulatedresults}
\centerline{\includegraphics[width=.5\textwidth]{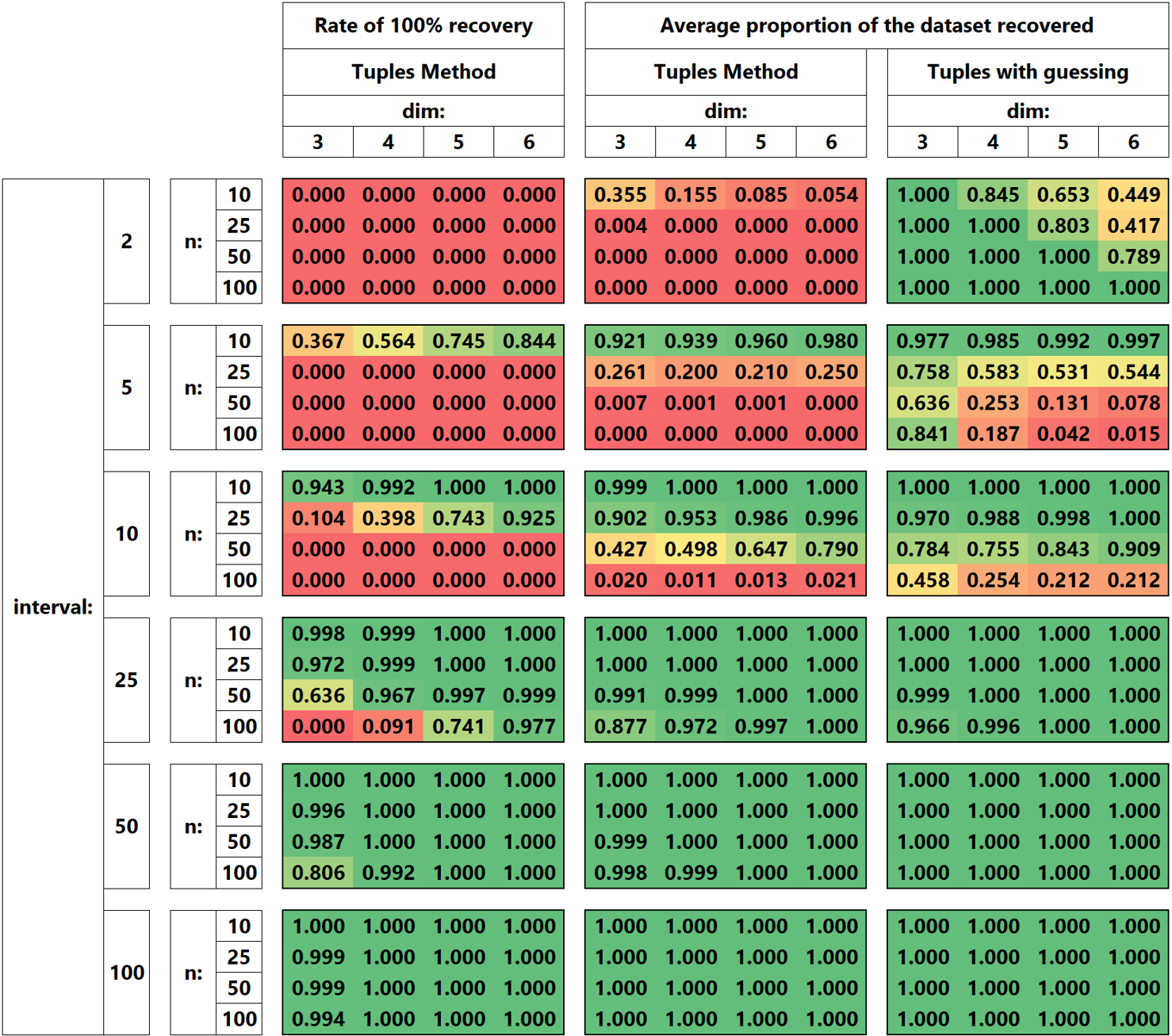}}
\end{table}

Since we were unable to develop a general formula to describe the expectation for each of these metrics for some set of input parameters, we ran a set of Monte Carlo simulations to get a sense of what the results looked like. The results in Table \ref{simulatedresults} represent many simulations of datasets with different parameters for dimension, n, and uniform interval. Randomly generated datasets were evaluated on each of the metrics discussed. All proportions calculated to judge the Tuples Method alone reflect 2000 simulations each, while those that judge the additional inferred likelihood methods reflect 100 simulations each (due to the increased computational complexity introduced).

In general, these results tell us that datasets will be recovered more effectively and completely when they have a relatively high dimension, a relatively low n, and a relatively high interval. All of these are directional indicators of how to reduce the number of phantoms found. With every added dimension, the cliques that are being searched for in the graph representation of the bivariate projections become more complex and are harder to happen by accident. With a lower n, there are fewer opportunities for overlapping edges to create phantoms accidentally – a sparser graph is easier to deal with and is less messy than a dense one. Finally, a higher interval means a higher chance of there being unique values within the columns of the dataset, which increases the odds that the Tuples Method will be more successful. Note that the Tuples Method, with inferred likelihood, eventually gets more successful with greater n. This is because with high enough n, there are enough duplicate rows that there is a greater chance of random success, as there are a limited number of possible rows for some dimension and interval in the first place.

\subsection{Results on Real-World Data}

As has been previously noted, Table \ref{simulatedresults} shows the results of simulations of these methods applied to randomly generated datasets within some set of parameters. A reasonable question one might ask is: how will the methods perform when used on real-world datasets? What will happen when variables are correlated with each other? To test this, we selected three real datasets to see how well they would be recovered by these methods. These results are shown in Table \ref{realresults}.

\begin{table}[htbp]
\caption{Real-World Results}
\begin{center}
\begin{tabular}{ |c|c|c|c|c|c|c| }
    \hline
    Dataset & dim & n & Cliques & Deduced & Expected & Actual \\
    \hline
    Hardware & 6 & 209 & 578 & 102 & 21.8\% & 77.5\% \\
    Iris & 5 & 150 & 321 & 63 & 28.8\% & 71.3\% \\
    Facebook 1 & 4 & 426 & 1621 & 9 & 7.2\% & 22.3\% \\
    Facebook 2 & 5 & 426 & 11819 & 23 & 0.3\% & 17.1\% \\
    \hline
\end{tabular}
\label{realresults}
\end{center}
\end{table}

We used three datasets from the UCI Machine Learning Repository: Computer Hardware, Iris, and Facebook.\footnote{For the Hardware dataset, we used columns 3-8. For the Iris dataset, all five columns were used. For the Facebook dataset, two versions were tested: one filtered on photos only with just columns 4-7 (“Facebook 1” in the table), and another (also filtered on photos) with columns 4-7 as well as column 16 (“Facebook 2” in the table).} In Table \ref{realresults}, “dim” refers to the dataset’s dimensionality, while “n” is the number of rows. The intervals of the datasets are difficult to quantify, since each variable may have a different interval, and it may not be possible to easily determine the total number of possibilities for a given variable (and in some cases, there is no limit, as, for example, values can get arbitrarily high).\footnote{So, as a proxy, the number of distinct values in each column is as follows: for Machine, there are 60, 25, 23, 22, 15, and 31; for Iris, there are 35, 23, 43, 22, and 3; and for Facebook, there are 12, 7, 22, and 2 (and 42 for the version with comments).}

“Cliques” refers to the total number of possible rows found from the graph representation of the dataset’s bivariate projections, and in each case, there were phantoms among the real cliques. “Deduced” refers to the number of rows that were deduced by the Tuples Method, and then “Expected” refers to the expected proportion of the dataset to be recovered correctly if, after the Tuples Method has deduced some number of rows, the remaining spaces were to be filled in by random guessing. Finally, “Actual” represents the actual proportion of the distinct rows of the original dataset that were in the reconstruction following the implementation of the inferred likelihood methods, which should ideally be greater than what you would expect from random guessing. We can see that in each case, the value of “Actual” is significantly higher than the value of “Expected,” implying success.

Though this is, of course, a small sample size, we do observe that the datasets with higher dim, lower n, and higher interval are recovered at a greater rate than datasets with the opposite characteristics, which is in alignment with the theoretical results gleaned from the simulated random examples. It should be noted that these methods do assume that the original number of rows of the dataset will be known, and this may not always be the case.


\begin{figure}[htbp]
\centering
\includegraphics[width=3.5in]{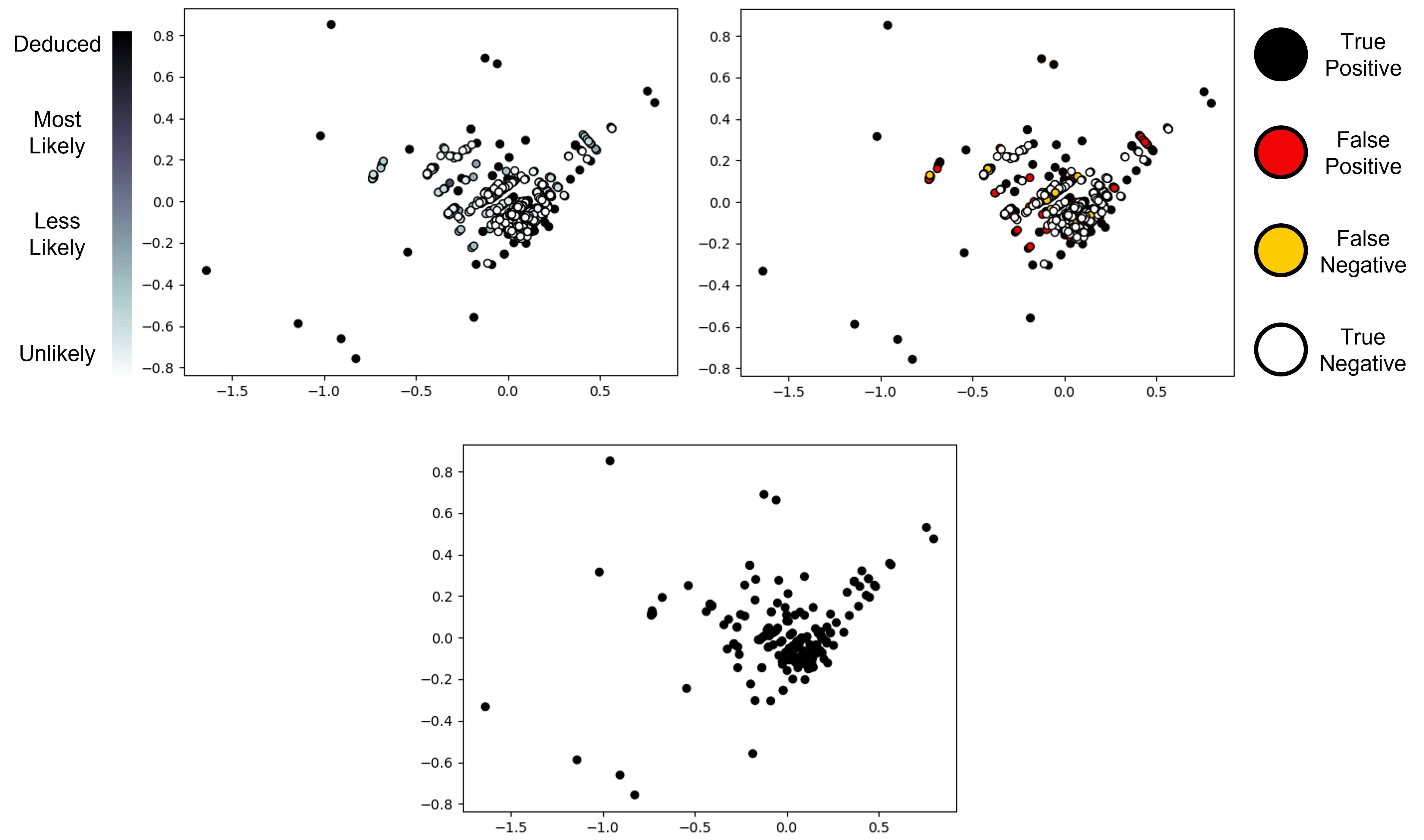}
\caption{Likelihood (left), accuracy (right), and real (bottom) MDS plots of Hardware dataset}
\label{hardwaremds}
\end{figure}

\begin{figure}[htbp]
\centering
\includegraphics[width=3.5in]{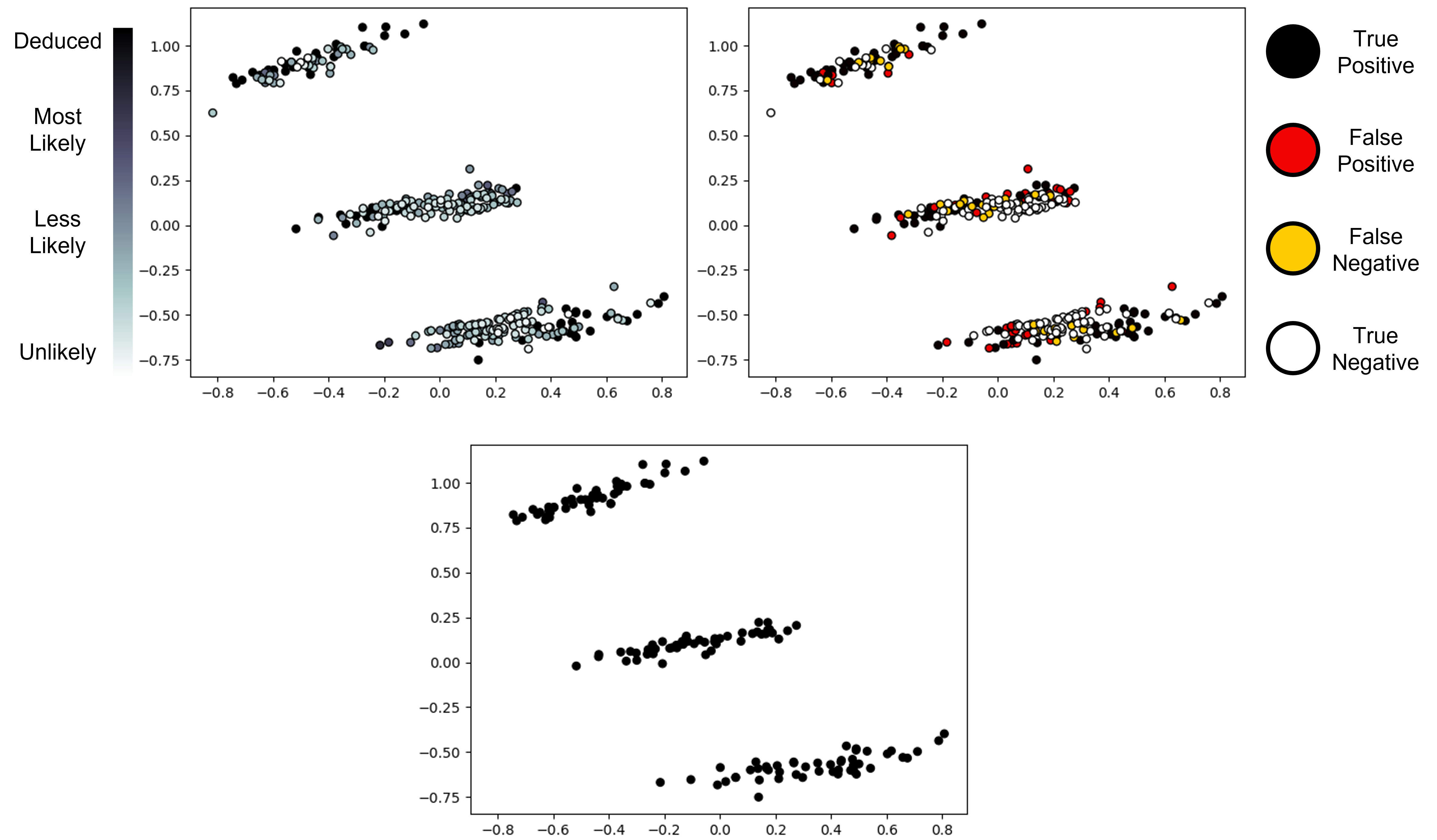}
\caption{Likelihood (left), accuracy (right), and real (bottom) MDS plots of Iris dataset}
\label{irismds}
\end{figure}

\begin{figure}[htbp]
\centering
\includegraphics[width=3.5in]{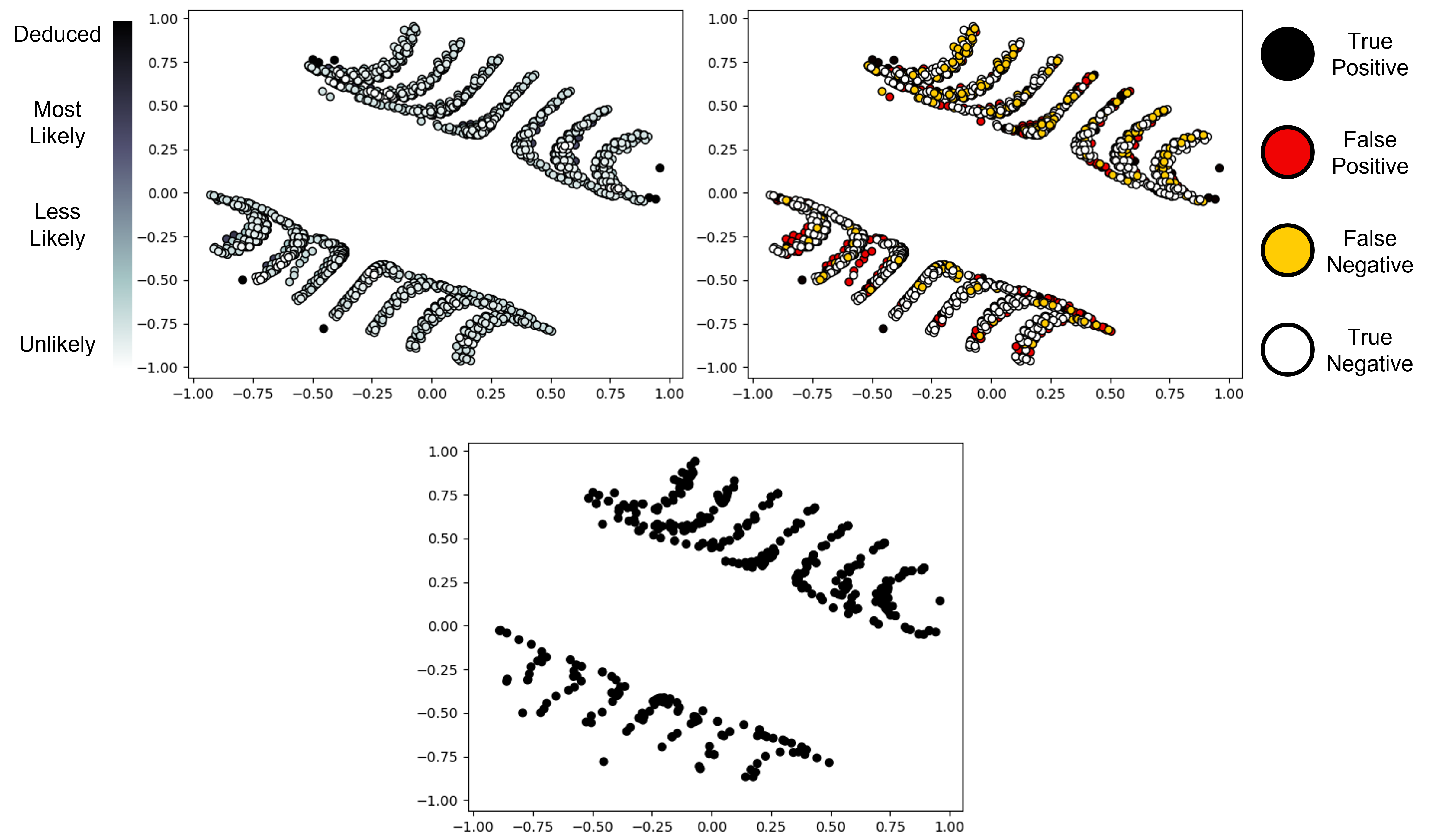}
\caption{Likelihood (left), accuracy (right), and real (bottom) MDS plots of Facebook 1 dataset}
\label{facebookmds}
\end{figure}

\begin{figure}[htbp]
\centering
\includegraphics[width=3.5in]{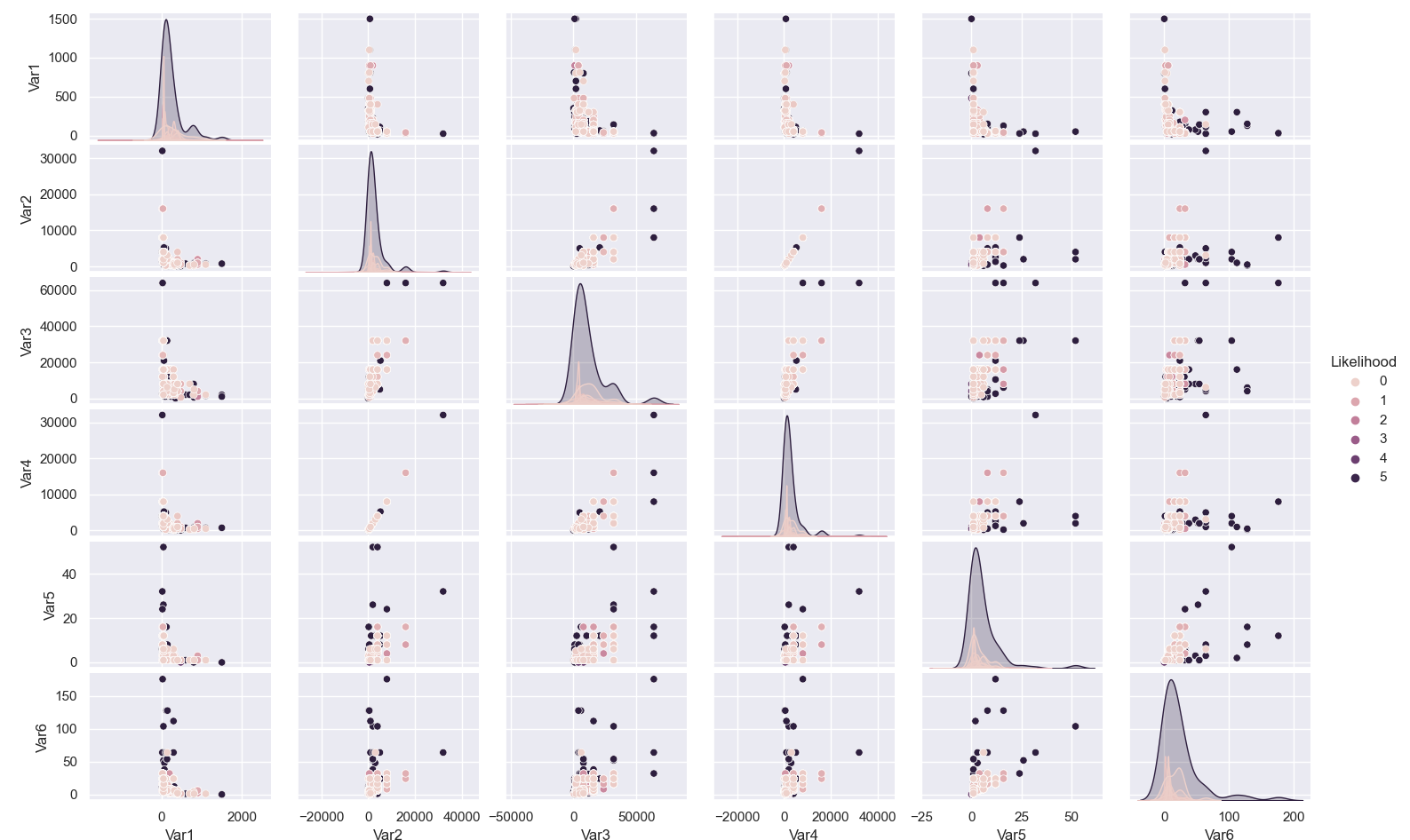}
\caption{Scatterplot matrix of Hardware dataset}
\label{hardwarespm}
\end{figure}

\begin{figure}[htbp]
\centering
\includegraphics[width=3.5in]{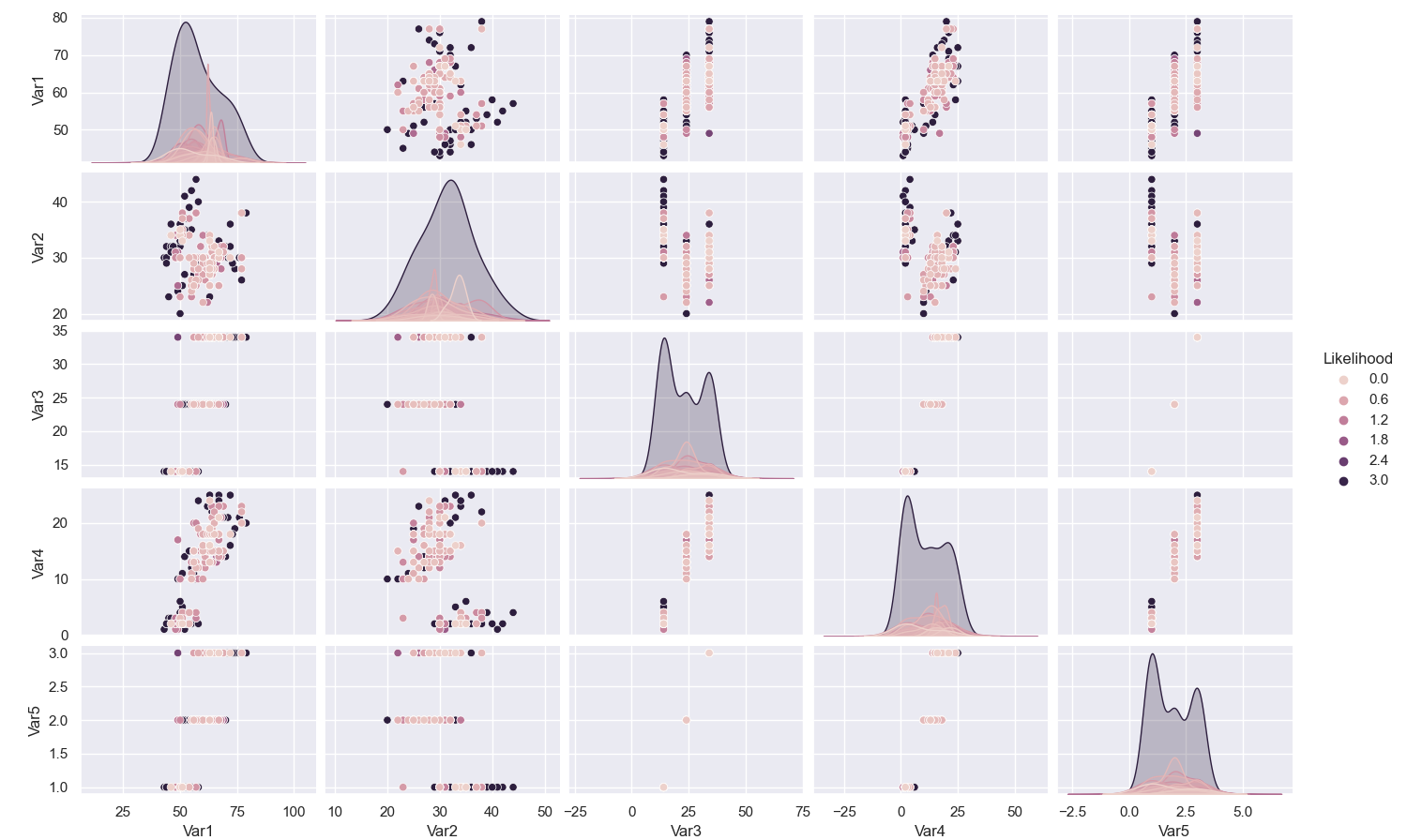}
\caption{Scatterplot matrix of Iris dataset}
\label{irisspm}
\end{figure}

\begin{figure}[htbp]
\centering
\includegraphics[width=3.5in]{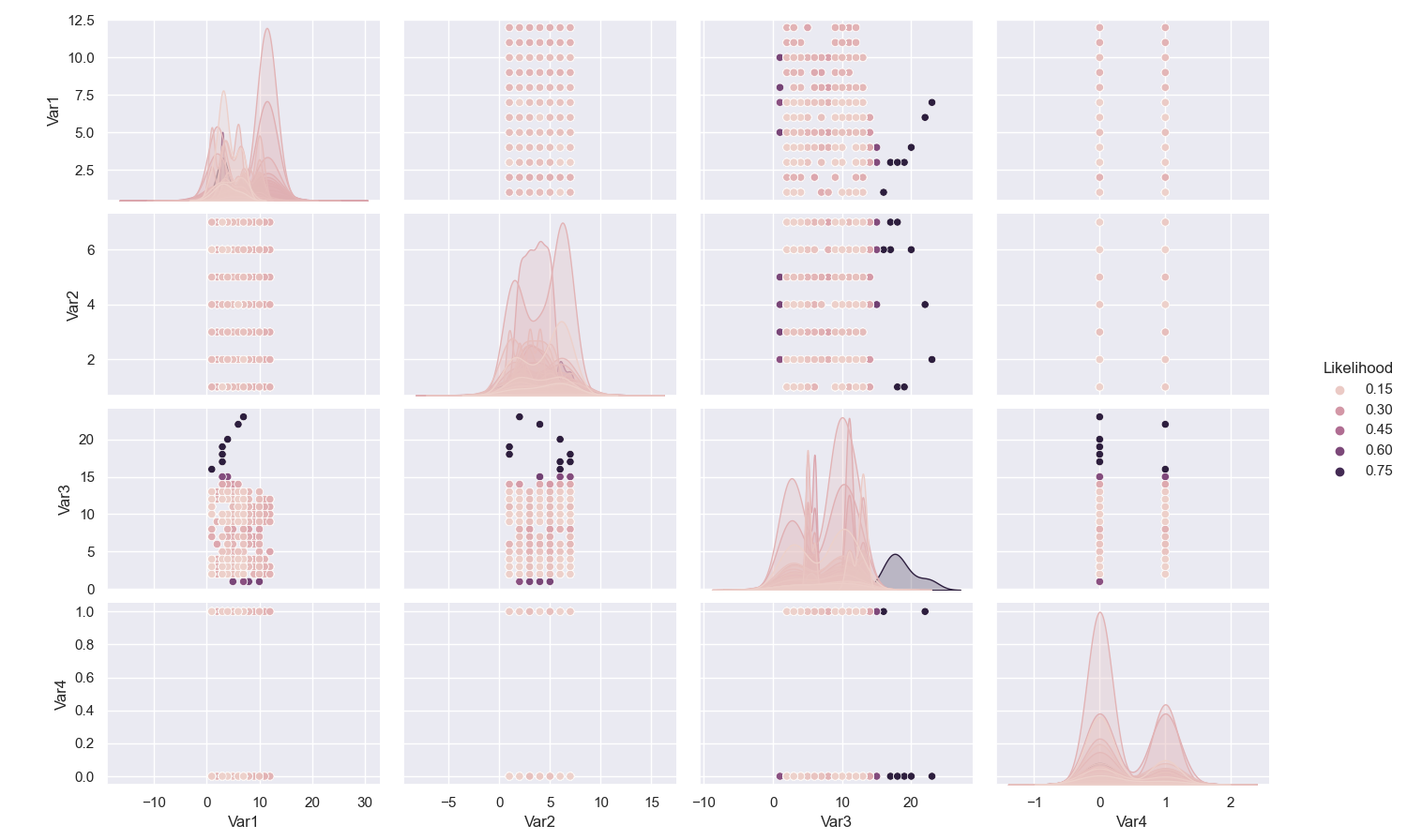}
\caption{Scatterplot matrix of Facebook dataset}
\label{facebookspm}
\end{figure}

For each of the datasets (other than Facebook 2, due to its size), we can visually display the process of reconstructing the dataset in a similar way as was shown in Section \ref{pcasection}. For this demonstration, instead of PCA, we chose to use multidimensional scaling (MDS) to reduce the dimensionality of the dataset. In each of the top two plots in Figs. \ref{hardwaremds}, \ref{irismds}, and \ref{facebookmds}, each point still represents each found clique. On the top left in each, we see the recovered rows in black and all other possibilities on a gradient reflecting their likelihood, and on the top right, we see the accuracy of the selections that were made. The bottom plot in each figure shows all of the rows that were in the original dataset. The scatterplot matrices for each of these datasets are given in Figs. \ref{hardwarespm}, \ref{irisspm}, and \ref{facebookspm}.

In these MDS plots, we get a sense of how these methods work on more structured data. We can see that the data itself tends to cluster according to internal patterns, but mini clusters also tend to emerge within the larger patterns. What happens here is that having a few rows that are close together in MDS space can actually cause many phantoms to come about, each of which borrows heavily from those few original rows, so they tend to cluster when visualized in this space. The points that ended up being correct tend to be spaced out and have “representation” in each substructure, since there is generally some original row to enable that substructure to exist in the first place. Finally, the points in these plots that are further away from other points are more often real; they also tend to be the ones that are more easily deduced since there are not as many possibilities that they are “close” to.

\section{Conclusion}

In summary, we have put forward several methods to solve the question of recovering datasets from their bivariate projections. Though we showed that complete recovery will not be possible in all cases, our given methods differentiate between exact deduction and when we are inferring which of several possibilities are most likely.

There are several possible next steps to continue working in this space, and we think that the top one would be to identify formulas for failure rates of methods beyond the Lookup Method. This would allow for accurate projections of how many phantoms are expected, the expected success of the Tuples Method, and the expected proportion of the original dataset that we might expect to recover on average for some given set of parameters. Ideally, such formulas could consider different intervals for different variables, as well as possibly having some understanding of different ways that the data can be distributed (and perhaps also correlations between given variables). Additionally, there is a possibility to expand the methods of deduction further beyond the Tuples Method, to minimize the need for methods of inferred likelihood.

More research could be done to generalize these methods more formally beyond bivariate projections into projections of arbitrary dimension (especially in generalizing the Tuples Method), as well as determining how best to account for information concerning the required number of specific edges that are yet to be accounted for in the inferred likelihood process. Also, we believe that these theories could potentially lead in interesting directions in the fields of error detection and error correction by building in some degree of redundancy in the data that can be reconstructed by the receiver of the message. Further, though it is conceptually possible with the present work, it would be interesting to see an implementation of these methods that allows a user to factor in their opinion to select rows of the original dataset when ambiguous cases do arise (these human selections could be considered as “recovered” and the inferred likelihood process could be run again with this new information). A few potential extensions that could lead to more interesting results would be investigating how this question fits into the Set Cover Problem, seeing what underlying group theoretic structures there may be here, and also connecting this to the problem of solving overdetermined linear systems.

Finally, in this work, we have assumed that the coordinates of the points in the bivariate projections have been given, and this is not always the case. Sometimes all one has is an image of a scatterplot matrix. To capture the point coordinates from these images, one could use the method by Cliche et al. \cite{cliche2017scatteract} or the well-known YOLO method by Redmon et al. \cite{redmon2016you}. We plan to experiment with these methods in future work.

\section*{Acknowledgments}
This research was supported by the Data + Computing = Discovery! (DCD) REU site, sponsored by NSF grant 1950052. Special thanks to Matthew Reuter, Alan Calder, and Chelsey Dollinger for running the DCD program, as well as to all other DCD participants. The visuals in the paper were created using Excel, PowerPoint, Matplotlib, and the graph editor on csacademy.com. The real-world datasets were downloaded from data.world and were originally uploaded there by UCI.

\bibliographystyle{IEEEtran}
\bibliography{bibliography}

\begin{IEEEbiography}[{\includegraphics[width=1in,height=1.25in,clip,keepaspectratio]{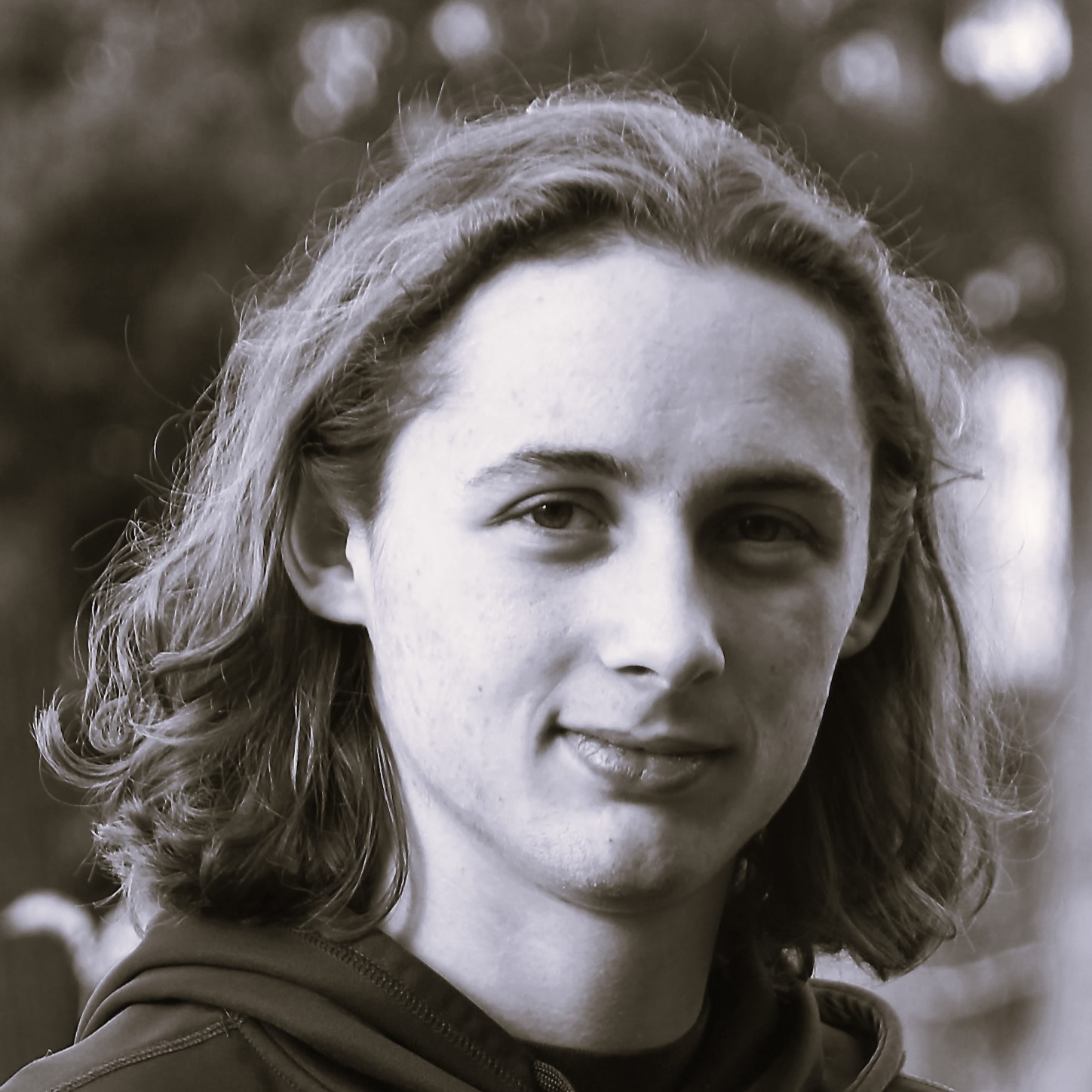}}]
{Eli Dugan} is a current sophomore at Williams College majoring in mathematics and philosophy. After graduation, he plans to pursue a PhD in mathematics, and a career in research and academia as a professor. He is currently serving on the board of the Williams Undergraduate Research Journal. He also works as both a TA and tutor in various math classes, providing one-on-one support to help students improve their understanding and confidence in the subject.
\end{IEEEbiography}

\begin{IEEEbiography}[{\includegraphics[width=1in,height=1.25in,clip,keepaspectratio]{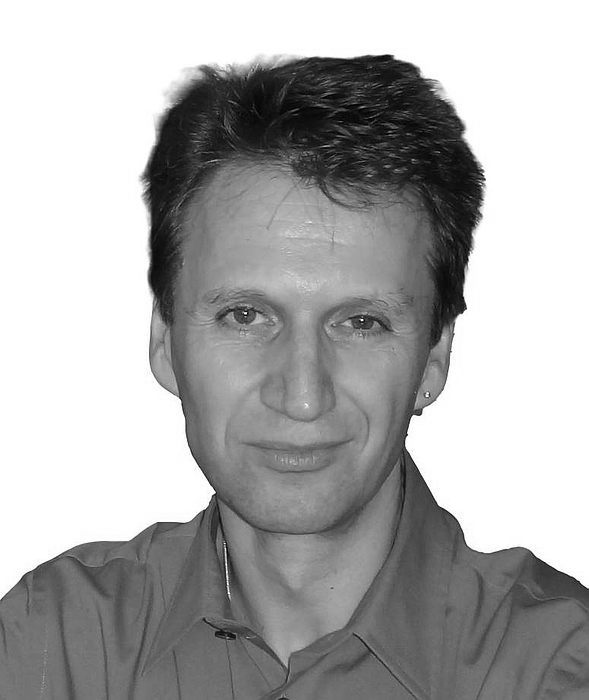}}]
{Klaus Mueller} has a PhD in computer science and is  currently  a  professor  of  computer  science  at  Stony  Brook  University  and  a  senior scientist at Brookhaven National Lab. His current research interests include explainable AI, visual analytics, data science, and medical imaging. He won the US National Science Foundation Early CAREER Award, the SUNY Chancellor’s Award for Excellence in Scholarship \& Creative Activity, and the IEEE CS Meritorious Service Certificate. His 300+ papers were cited over 12,500 times.
\end{IEEEbiography}

\end{document}